\documentclass[11pt]{article}
\usepackage[utf8]{inputenc}
\usepackage{acl2016}
\usepackage{times}
\usepackage{latexsym}
\usepackage{amsfonts}
\usepackage{url}

\aclfinalcopy 

\title{Sentence Pair Scoring: Towards Unified Framework \\ for Text Comprehension}

\author{Petr Baudi\v{s},
	Jan Pichl,
	Tom\'{a}\v{s} Vysko\v{c}il
	\and
	Jan \v{S}ediv\'{y} \\
	FEE CTU Prague\\
	Department of Cybernetics\\
	Technick\'{a} 2, Prague, Czech Republic\\
	{\tt baudipet@fel.cvut.cz}}

\date{}

\begin{document}

\maketitle

\begin{abstract}
We review the task of Sentence Pair Scoring, popular in the literature
in various forms --- viewed as Answer Sentence Selection,
Semantic Text Scoring, Next Utterance Ranking, Recognizing Textual Entailment, Paraphrasing
or e.g.\ a component of Memory Networks.

We argue that all such tasks are similar from the model perspective and propose new baselines by
comparing the performance of common IR metrics and popular convolutional, recurrent and attention-based
neural models across many Sentence Pair Scoring tasks and datasets.
We discuss the problem of evaluating randomized models, propose a statistically
grounded methodology, and attempt to improve comparisons by releasing new
datasets that are much harder than some of the currently used well explored
benchmarks.
%
We introduce a unified open source software framework with easily pluggable
models and tasks, which enables us to experiment with
multi-task reusability of trained sentence models.
We set a new state-of-art in performance on the Ubuntu Dialogue dataset.
\end{abstract}

\section{Introduction}

An NLP machine learning task often involves classifying a sequence of tokens
such as a sentence or a document, i.e.\ approximating a function
	$f_1(s) \in [0,1]$
(where $f_1$ may determine a domain, sentiment, etc.).  But there is a large
class of problems that involve classifying a pair of sentences,
	$f_2(s_0, s_1) \in \mathbb{R}$
(where $s_0$, $s_1$ are sequences of tokens, typically sentences).

Typically, the function $f_2$ represents some sort of \textit{semantic similarity},
that is whether (or how much) the two sequences are semantically related.
This formulation allows $f_2$ to be a measure for tasks as different as topic relatedness,
paraphrasing, degree of entailment, a pointwise ranking task for answer-bearing
sentences or next utterance classification.

In this work, we adopt the working assumption that there exist certain universal
$f_2$ type measures that may be successfuly applied to a wide variety of semantic
similarity tasks --- in the case of neural network models trained to represent universal semantic comprehension of sentences
and adapted to the given task by just fine-tuning or adapting the output neural layer
(in terms of architecture or just weights).
Our argument for preferring $f_2$ to $f_1$ in this pursuit is the fact that the other
sentence in the pair is essentially a very complex label when training the sequence model,
which can therefore discern semantically rich structures and dependencies.
Determining and demonstrating such universal semantic comprehension models
across multiple tasks remains a few steps ahead, since the research landscape
is fragmented in this regard.  Model research is typically reported within the
context of just a single $f_2$-type task, each dataset requires sometimes
substantial engineering work before measurements are possible, and results
are reported in ways that make meaningful model comparisons problematic.

Our main aims are as follows. (A) Unify research within a single framework that employs
task-independent models and task-specific adaptation modules.  (B) Improve the methodology
of model evaluation in terms of statistics, comparing with strong non-neural
IR baselines, and introducing new datasets with better characteristics.
(C) Demonstrate the feasibility
of pursuing universal, task-independent $f_2$ models, showing that even simple
neural models learn universal semantic comprehension by employing cross-task transfer
learning.

The paper is structured as follows.  In Sec.~\ref{sec:tasks}, we outline possible
specific $f_2$ tasks and available datasets; in Sec.~\ref{sec:models}, we survey
the popular non-neural and neural baselines in the context of these tasks;
finally, in Sec.~\ref{sec:perf}, we present model-task evaluations within
a unified framework to establish the watermark for future research as well as gain
insight into the suitability of models across a variety of tasks.  In
Sec.~\ref{sec:transfer}, we demonstrate that transfer learning across tasks is helpful
to powerfully seed models. 
We conclude with Sec.~\ref{sec:concl}, summarizing our findings and outlining several
future research directions.

\section{Tasks and Datasets}
\label{sec:tasks}

The tasks we are aware of that can be phrased as $f_2$-type problems are listed below.
In general, we primarily focus on tasks that have reasonably large and
realistically complex datasets freely available.  On the contrary, we have explicitly
avoided datasets that have licence restrictions on availability or commercial usage.

\subsection{Answer Sentence Selection}

Given a factoid question and a set of candidate answer-bearing sentences in encyclopedic style,
the first task is to rank higher sentences that are more likely to contain the answer to the question.
As it is fundamentally an Information Retrival task in nature, the model performance
is commonly evaluated in terms of Mean Average Precision (MAP) and Mean Reciprocial
Rank (MRR).

This task is popular in the NLP research community thanks to the dataset
introduced in \cite{AnsselWang} (which we refer to as \texttt{wang}),
with six papers published between February 2015 and 2016 alone
and neural models substantially improving over classical approaches based primarily on parse tree edits.%
\footnote{\url{http://aclweb.org/aclwiki/index.php?title=Question_Answering_(State_of_the_art)}}
It is possibly the main research testbed for $f_2$-style task models.
This task has also immediate applications e.g.\ in Question Answering systems.

In the context of practical applications, the so-far standard \texttt{wang} dataset
has several downsides we observed when tuning and evaluating our models,
illustrated numerically in Fig.~\ref{tab:ansseldata} --- the set of candidate sentences is often very small
and quite uneven (which also makes rank-based measures unstable) and the total
number of individual sentence pairs as well as questions is relatively small.
Furthermore, the validation and test set are very small, which makes for noisy
performance measurements; the splits also seem quite different in the nature of questions
since we see minimum correlation between performance on the validation and test sets,
which calls the parameter tuning procedures and epoch selection for early stopping
into question.
Alternative datasets WikiQA \cite{WikiQA} and InsuranceQA \cite{attn1511} were proposed, but are encumbered
by licence restrictions.  Furthermore, we speculate that they may suffer from many of the problems above%
\footnote{Moreover, InsuranceQA is effectively a classification task rather than a ranking task,
	which we do not find as appealing in the context of practical applications.}
(even if they are somewhat larger).

To alleviate the problems listed above, we are introducing a new dataset
\texttt{yodaqa/large2470} based on an extension of the \texttt{curatedv2} question
dataset (introduced in \cite{YodaQACLEF2015}, further denoisified by Mechanical Turkers)
with candidate sentences as retrieved by the YodaQA question answering system \cite{YodaQAPoster2015}
from English Wikipedia and labelled by matching the gold standard answers in the passages.%
\footnote{Note that the \texttt{wang} and \texttt{yodaqa} datasets however
share a common ancestry regarding the set of questions and there may be some
overlaps, even across train and test splits.  Therefore, mixing training and
evaluation on wang and yodaqa datasets within a single model instance is not advisable.}

Motivated by another problem related to the YodaQA system, we also introduce
another dataset \texttt{wqmprop}, where $s_0$ are again question sentences,
but $s_1$ are English labels of properties that make a path within the Freebase
knowledge base that connects an entity linked in the question to the correct
answer.  This task \textbf{(Property Selection)} can be evaluated identically
to the previous task, and solutions often involving Convolutional Neural Networks
have been studied in the Question Answering literature \cite{STAGG} \cite{qatextev}.
Our sentences have been derived from the WebQuestions dataset \cite{WebQuestions}
extended with the moviesE dataset questions (originally introduced in \cite{YodaQACLEF2015});
the property paths are based on the Freebase knowledge graph dump,
generated based on entity linking and exploration procedure of YodaQA v1.5.%
\footnote{\url{https://github.com/brmson/dataset-factoid-webquestions} branch \textit{movies}}

Fig.~\ref{tab:ansseldata} compares the critical characteristics of the datasets.
Furthermore, as apparent below, the baseline performances on the newly proposed
datasets are much lower, which suggests that future model improvements will be more
apparent in evaluation.

\begin{figure*}[t]
\centering
\begin{tabular}{|c|ccc|c|cc|}
\hline
Dataset & Train pairs & Val. pairs & Test pairs & Val.-Test $r$ & Ev. $\#s_0$ & Ev. $\#s_1$ per $s_0$ \\
\hline
\texttt{wang} & 44648 & 1149 & 1518 & -0.078 & 178 & 34.9 $\pm$131\% \\
\texttt{yodaqa/large2470} & 220846 & 55052 & 120069 & 0.348 & 1100 & 159.2 $\pm$100\% \\
\hline
\texttt{wqmprop} & 407465 & 137235 & 277509 & 0.836 & 3430 & 118.753 $\pm$85\% \\
\hline
Ubuntu Dialogue v2 & 1M & 195600 & 189200 & 0.884 & 38480 & 10 \\
\hline
\end{tabular}
\vspace*{-0.2cm}
\caption{\footnotesize%
The Val.-Test column shows inter-trial Pearson's $r$ of validation and test MRRs,
averaged across the models we benchmarked (see below).
The $s_0$ and $s_1$ statistics are shown for the evaluation (Ev. --- validation and test)
portion of the datasets.
The last column includes
relative standard deviation of the number of candidate sentences per question, which corresponds
to the variation in the difficulty of the ranking task (as well as variation in expected
measure values for individual questions).
}
\label{tab:ansseldata}
\end{figure*}

\subsection{Next Utterance Ranking}

\cite{UbuntuLowe} proposed the new large-scale real-world Ubuntu Dialogue dataset for an $f_2$-style
task of ranking candidates for the next utterance in a chat dialog, given the dialog
context.  The technical formulation of the task is the same as for Answer Sentence Selection,
but semantically, choosing the best followup has different concerns than
choosing an answer-bearing sentence.
Recall at top-ranked 1, 2 or 5 utterances out of either 2 or 10 candidates is reported;
we also propose reporting the utterance MRR as a more aggregate measure.
The newly proposed Ubuntu Dialogue dataset is based on IRC chat logs of the Ubuntu
community technical support channels and contains casually typed interactions regarding
computer-related problems.\footnote{In a manner, they resemble tweet data, but without the length restriction and with heavily technical jargon, interspersed command sequences etc.}
While the training set consists of individual labelled pairs,
during evaluation 10 followups to given message(s) are ranked.  The sequences might be over 200 tokens long.

Our primary motivation for using this dataset is its size.
The numerical characteristics
of this dataset are shown in Table~\ref{tab:ansseldata}.%
\footnote{As in past papers, we use only the first 1M pairs (10\%) of the training set.}
We use the v2 version of the dataset.%
\footnote{\url{https://github.com/rkadlec/ubuntu-ranking-dataset-creator}}
Research published on this dataset so far relies on simple neural models.
\cite{UbuntuLowe} \cite{UbuntuKadlec}

\subsection{Recognizing Textual Entailment and Semantic Textual Similarity}

One of the classic tasks at the boundary of Natural Language Processing
and Artificial Intelligence is the inference problem of Recognizing Textual
Entailment \cite{RTE1} --- given a pair of a factual sentence and a hypothesis
sentence, we are to determine whether the hypothesis represents a contradiction,
entailment or is neutral (cannot be proven or disproven).

We include two current popular machine learning datasets for this task.
The Stanford Natural Language Inference \textbf{SNLI} dataset \cite{SNLI} consists
of 570k English sentence pairs with the facts based on image captions,
and 10k + 10k of the pairs held out as validation and test sets.
The \textbf{SICK-2014} dataset \cite{SICK2014}
was introduced as Task 1 of the SemEval 2014 conference and in contrast to SNLI,
it is geared at specifically benchmarking semantic compositional methods, aiming to
capture only similarities on purely language and common knowledge level,
without relying on domain knowledge, and there are no named entities or
multi-word idioms; it consists of 4500 training pairs, 500 validation pairs
and 4927 testing pairs.

For the SICK-2014 dataset, we also report results on the Semantic Textual Similarity.
This task originates in the STS track of the
SemEval conferences \cite{STS2015} and involves scoring
pairs of sentences from 0 to 5 with the objective of maximizing correlation
(Pearson's $r$) with manually annotated gold standard.

\section{Models}
\label{sec:models}

As our goal is a universal text comprehension model,
we focus on neural network models architecture-wise.
We assume that the sequence is transformed using $N$-dimensional word embeddings
on input, and employ models that produce a pair of sentence embeddings $E_0$, $E_1$
from the sequences of word embeddings ${e_0}$, ${e_1}$.
Unless noted otherwise, a Siamese architecture is used that shares weights
among both sentenes.

A scorer module that compares the $E_0, E_1$ sentence embeddings to produce
a scalar result is connected to the model; for specific task-model configurations,
we use either the \textbf{dot-product}
module $E_0 \cdot E_1^T$ (representing non-normalized vector angle, as in e.g. \cite{Yu14} or \cite{MemNN})%
\footnote{Not normalizing the vectors acts as a regularization for their size.
In all our experiments, cosine distance fared much worse.}
or the
\textbf{MLP} module that takes elementwise product and sum of the embeddings
and feeds them to a two-layer perceptron with hidden layer of width $2N$
(as in e.g. \cite{TreeLSTM}).%
\footnote{The motivation is to capture both angle and euclid distance
in multiple weighed sums.  Past literature uses absolute difference rather
than sum, but both performed equally in our experiments and we adopted
sum for technical reasons.}
For the STS task, we follow this by score regression using class interpolation as in \cite{TreeLSTM}.

When training for a ranking task (Answer Sentence Selection), we use the bipartite
ranking version of Ranknet \cite{Ranknet} as the objective; when training for STS task,
we use Pearson's $r$ formula as the objective; for binary classification tasks, we use
the binary crossentropy objective.

\subsection{Baselines}

In order to anchor the reported performance, we report several basic methods.
\textbf{Weighed word overlaps} metrics TF-IDF and BM25 \cite{BM25} are inspired
by IR research and provide strong baselines for many tasks.  We treat $s_0$ as
the query and $s_1$ as the document, counting the number of common words and
weighing them appropriately.  IDF is determined on the training set.

The \textbf{avg} metric represents the baseline method when using word embeddings
that proved successful e.g.\ in \cite{Yu14} or \cite{MemNN}, simply
taking the mean vector of the word embedding sequence and training an $U$ weight matrix
$N\times 2N$
that projects both embeddings to the same vector space,
$E_i = \tanh(U\cdot\bar{e_i})$,
where the MLP scorer compares them.  During training, $p=1/3$ standard (elementwise) dropout
is applied on the input embeddings.

A simple extension of the above are the \textbf{DAN} Deep Averaging Networks
\cite{DAN}, which were shown to adequately replace much more complex models in some
tasks.  Two dense perceptron layers are stacked between the mean and projection,
relu is used instead of tanh as the non-linearity, and word-level dropout is
used instead of elementwise dropout.

\subsection{Recurrent Neural Networks}

\textbf{RNN} with memory units are popular models for processing
sentenes \cite{attn1511} \cite{UbuntuLowe} \cite{SNLI}.  We use a bidirectional
network with $2N$ GRU memory units%
\footnote{While the LSTM architecture is more popular, we have found the GRU results
are equivalent while the number of parameters is reduced.} \cite{GRU}
in each direction; the final unit states are summed across the per-direction GRUs
to yield a $2N$ vector representation of the sentence.
Like in the avg baseline, a projection matrix is applied on this representation
and final vectors compared by an MLP scorer.
We have found that applying massive dropout $p=4/5$ both on the input and output
of the network helps to avoid overfitting even early in the training.

\subsection{Convolutional Neural Networks}

\textbf{CNN} with sentence-wide pooling layer are also popular models for processing
sentences \cite{Yu14} \cite{attn1511} \cite{SevMos2015} \cite{MuPeCNN} \cite{UbuntuKadlec}.
We apply a multi-channel convolution \cite{KimMultichannelCNN} with single-token channel
of $N$ convolutions and 2, 3, 4 and 5-token channels of $N/2$ convolutions each, relu transfer
function, max-pooling over the whole sentence, and as above a projection to shared space
and an MLP scorer.  Dropout is not applied.

\subsection{RNN-CNN Model}

The \textbf{RNN-CNN} model aims to combine both recurrent and convolutional networks
by using the memory unit states in each token as the new representation of the token
which is then fed to the convolutional network.  Inspired by \cite{attn1511}, the aim
of this model is to allow the RNN to model long-term dependencies and model contextual
representations of words, while taking advantage of the CNN and pooling operation for
crisp selection of the gist of the sentence.  We use the same parameters as for the
individual models, but with no dropout and reducing
the number of parameters by using only $N$ memory units per direction.

\subsection{Attention-Based Models}

The idea of attention models is to attend preferrentially to some parts of the sentence
when building its representation \cite{ReadComprehend} \cite{attn1511} \cite{attnpooling} \cite{SNLIattn}.
There are many ways to model attention, we adopt the
\cite{attn1511} model \textbf{attn1511} as a conceptually simple and easy to implement baseline.
It asymmetrically extends the RNN-CNN model by extra links from $s_0$ CNN output to
the post-recurrent representation of each $s_1$ token, determining an attention level
for each token by weighed sum of the token vector elements, focusing on the relevant $s_1$
segment by transforming the attention levels using softmax and multiplying the token
representations by the attention levels before they are fed to the convolutional network.

Convolutional network weights are not shared between the two sentences and the convolutional
network output is not projected before applying the MLP scorer.  The CNN used here
is single-channel with $2N$ convolution filters 3 tokens wide.

\section{Model Performance}
\label{sec:perf}

\subsection{\texttt{dataset-sts} framework}

To easily implement models, dataset loaders and task adapters in a modular fashion
so that any model can be easily run on any $f_2$-type task, we have created
a new software package \texttt{dataset-sts} that integrates a variety of datasets,
a Python dataset adapter \texttt{PySTS} and a Python library for easy construction
of deep neural NLP models for semantic sentence pair scoring \texttt{KeraSTS} that
uses the Keras machine learning library \cite{Keras}.  The framework is available
for other researchers as open source on GitHub.%
\footnote{\url{https://github.com/brmson/dataset-sts}}

\subsection{Experimental Setting}

We use $N=300$ dimensional GloVe embeddings matrix pretrained on Wikipedia 2014 + Gigaword 5 \cite{GloVe}
that we keep adaptable during training; words in the training set not included in the
pretrained model are initialized by random vectors uniformly sampled from $[-0.25, +0.25]$ to match
the embedding standard deviation.

Word overlap is an important feature in many $f_2$-type tasks \cite{Yu14} \cite{SevMos2015},
especially when the sentences may contain named entities, numeric or other data for which
no embedding is available.  As a workaround, ensemble of world overlap count and neural
model score is typically used to produce the final score.  In line with this idea,
in the Answer Sentence Selection
\texttt{wang} and \texttt{large2470} datasets, we use the BM25 overlap baseline
as an additional input to the MLP scoring module, and prune the scored samples to top 20
based on BM25.%
\footnote{This reduces the number of (massively irrelevant) training samples, but we observed no adverse effects of that, while it speeds up training greatly and models well a typical Information Retrieval scenario where fast pre-scoring of candidates is essential.}
Furthermore, we extend the embedding of each input token by several extra dimensions carrying boolean
flags --- bigram overlap, unigram overlap (except stopwords and interpunction), and whether
the token starts with a capital letter or is a number.

Particular hyperparameters are tuned primarily on the \texttt{yodaqa/large2470} dataset
unless noted otherwise in the respective results table caption.
We apply $10^{-4}$ $\mathbb{L}_2$ regularization and use Adam optimization with standard parameters \cite{Adam}.
In the answer selection tasks, we train on 1/4 of the dataset in each epoch.
After training, we use the epoch with best validation performance;
sadly, we typically observe heavy overfitting as training progresses and rarely use a model from later than a couple of epochs.

\subsection{Evaluation Methodology}

We report model performance averaged across 16 training runs (with
different seeds).  A consideration we must emphasize is that randomness
plays a large role in neural models both in terms of randomized weight
initialization and stochastic dropout.  For example, the typical methodology
for reporting results on the \texttt{wang} dataset is to evaluate and report
a single test run after tuning on the dev set,%
\footnote{Confirmed by personal communication with paper authors.}
but \texttt{wang} test MRR has empirical standard deviation of 0.025 across
repeated runs of our attn1511 model, which is more than twice the gap between every
two successive papers pushing the state-of-art on this dataset!
See the $^*$-marked sample in Fig.~\ref{tab:anssel} for a practical example of this phenomenon.
Furthermore, on more complex tasks
(Answer Sentence Selection in particular, see Fig.~\ref{tab:ansseldata}) the validation set performance
is not a great approximator for test set performance and a strategy like
picking the training run with best validation performance would lead just
to overfitting on the validation set.

To allow comparison between models (and with future models), we therefore
report also 95\% confidence intervals for each model performance estimate,
as determined from the empirical standard deviation using Student's
t-distribution.%
\footnote{Over larger number of samples, this estimate converges to the normal distribution confidence levels.
Note that the confidence interval determines the range of the true expected evaluation,
not evaluation of any measured sample.}

\begin{figure*}[t]
\centering
\setlength{\tabcolsep}{3pt}
\begin{tabular}{|c|cc|cc|cc|}
\hline
Model              & wang MAP & wang MRR & l2470 MAP & l2470 MRR & wqm MAP & wqm MRR \\
\hline
\cite{attn1511}    & 0.728 & 0.832 & & & & \\
\cite{attnpooling} & 0.753 & 0.851 & & & & \\
\hline
TF-IDF & $0.578$ & $0.709$ & $0.267$ & $0.363$ &  & \\
BM25 & $0.630$ & $0.765$ & $0.314$ & $0.491$ & $0.216$ & $0.194$\\
RNN w/o BM25 & $0.649$ & $0.743$ & $0.262$ & $0.381$ &  & \\
 & $\quad^{\pm0.011}$ & $\quad^{\pm0.010}$ & $\quad^{\pm0.003}$ & $\quad^{\pm0.008}$ &  & \\
\hline
avg & $0.713$ & $0.806$ & $0.278$ & $0.481$ & $0.462$ & $0.506$\\
 & $\quad^{\pm0.003}$ & $\quad^{\pm0.005}$ & $\quad^{\pm0.003}$ & $\quad^{\pm0.008}$ & $\quad^{\pm0.013}$ & $\quad^{\pm0.015}$\\
DAN & $0.709$ & $0.787$ & $0.282$ & $0.490$ & $0.457$ & $0.503$\\
 & $\quad^{\pm0.004}$ & $\quad^{\pm0.007}$ & $\quad^{\pm0.004}$ & $\quad^{\pm0.010}$ & $\quad^{\pm0.007}$ & $\quad^{\pm0.008}$\\
RNN & $0.696$ & $0.785$ & $0.277$ & $0.487$ & $0.653$ & $0.682$\\
 & $\quad^{\pm0.006}$ & $\quad^{\pm0.007}$ & $\quad^{\pm0.004}$ & $\quad^{\pm0.008}$ & $\quad^{\pm0.068}$ & $\quad^{\pm0.065}$\\
CNN & $0.717$ & $0.793$ & $0.288$ & $0.499$ & $0.664$ & $0.694$\\
 & $\quad^{\pm0.005}$ & $\quad^{\pm0.005}$ & $\quad^{\pm0.003}$ & $\quad^{\pm0.007}$ & $\quad^{\pm0.021}$ & $\quad^{\pm0.019}$\\
RNN-CNN & \textbf{0.729} & $0.810$ & $0.288$ & $0.503$ & $0.517$ & $0.556$\\
 & $\quad^{\pm0.006}$ & $\quad^{\pm0.009}$ & $\quad^{\pm0.004}$ & $\quad^{\pm0.010}$ & $\quad^{\pm0.052}$ & $\quad^{\pm0.048}$\\
attn1511 & \textbf{0.732} & $0.817$ & $0.286$ & $0.499$ & \textbf{0.701} & \textbf{0.729}\\
 & $\quad^{\pm0.006}$ & $\quad^{\pm0.012}$ & $\quad^{\pm0.003}$ & $\quad^{\pm0.009}$ & $\quad^{\pm0.008}$ & $\quad^{\pm0.005}$\\
$^*$attn1511       & $0.756$ & $0.859$ & & & & \\
\hline
Ubu. RNN w/o BM25 & \textbf{0.731} & $0.814$ & \textbf{0.359} & \textbf{0.539} &  & \\
 & $\quad^{\pm0.007}$ & $\quad^{\pm0.008}$ & $\quad^{\pm0.003}$ & $\quad^{\pm0.006}$ &  & \\
Ubu. RNN &  &  & $0.291$ & $0.515$ &  & \\
 &  &  & $\quad^{\pm0.002}$ & $\quad^{\pm0.004}$ &  & \\
SNLI RNN &  &  & $0.264$ & $0.460$ &  & \\
 &  &  & $\quad^{\pm0.005}$ & $\quad^{\pm0.015}$ &  & \\
\hline
\end{tabular}
\setlength{\tabcolsep}{6pt}
\vspace*{-0.2cm}
\caption{\footnotesize%
	Model results on the Answer Sentence Selection task,
	as measured on the \texttt{wang}, \texttt{yodaqa/large2470} and \texttt{wqmprop} datasets.
	\texttt{wqmprop} does not use the BM25 ensembling, and CNN is not Siamese.
	\\
	$^*$ Demonstration of the problematic single-measurement result reporting
	in past literature --- an outlier sample in our 16-trial attn1511 benchmark
	that would score as a state of art; in total, three outliers in the trial ($12.5\%$) scored better than \cite{attn1511}.
}
\label{tab:anssel}
\end{figure*}

\begin{figure*}[p]
\centering
\begin{tabular}{|c|c|c|ccc|}
\hline
Model & MRR & 1-2 R@1 & 1-10 R@1 & 1-10 R@2 & 1-10 R@5 \\
\hline
$^*$ TF-IDF & & 0.749 & 0.488 & 0.587 & 0.763 \\
$^*$ RNN & & 0.777 & 0.379 & 0.561 & 0.836 \\
$^*$ LSTM & & 0.869 & 0.552 & 0.721 & 0.924 \\
$^*$ MemN2N 3-hop & & & 0.637 & & \\
\hline
avg & 0.624 & 0.793 & 0.472 & 0.608 & 0.836 \\
 & $\quad^{\pm0.002}$ & $\quad^{\pm0.002}$ & $\quad^{\pm0.002}$ & $\quad^{\pm0.002}$ & $\quad^{\pm0.003}$ \\
DAN & 0.578 & 0.792 & 0.493 & 0.615 & 0.830 \\
 & $\quad^{\pm0.070}$ & $\quad^{\pm0.035}$ & $\quad^{\pm0.074}$ & $\quad^{\pm0.059}$ & $\quad^{\pm0.033}$ \\
\hline
RNN & 0.781 & 0.907 & 0.664 & 0.799 & 0.951 \\
 & $\quad^{\pm0.003}$ & $\quad^{\pm0.002}$ & $\quad^{\pm0.004}$ & $\quad^{\pm0.004}$ & $\quad^{\pm0.001}$ \\
CNN & 0.718 & 0.863 & 0.587 & 0.721 & 0.907 \\
 & $\quad^{\pm0.003}$ & $\quad^{\pm0.002}$ & $\quad^{\pm0.004}$ & $\quad^{\pm0.005}$ & $\quad^{\pm0.003}$ \\
RNN-CNN & \textbf{0.788} & \textbf{0.911} & \textbf{0.672} & \textbf{0.809} & \textbf{0.956} \\
 & $\quad^{\pm0.001}$ & $\quad^{\pm0.001}$ & $\quad^{\pm0.002}$ & $\quad^{\pm0.002}$ & $\quad^{\pm0.001}$ \\
attn1511 & 0.772 & 0.903 & 0.653 & 0.788 & 0.945 \\
 & $\quad^{\pm0.004}$ & $\quad^{\pm0.002}$ & $\quad^{\pm0.005}$ & $\quad^{\pm0.005}$ & $\quad^{\pm0.002}$ \\
\hline
\end{tabular}
\vspace*{-0.2cm}
\caption{\footnotesize%
	Model results on the Ubuntu Dialogue next utterance ranking task.
	Models use slightly specific configuration due to much bigger dataset (in terms of both samples and sentence lengths) --- only 160 tokens are considered per input, no dropout is applied,
	RNN use $N$ memory units, projection matrix is only $N\times N$ and the dot-product scorer is used for comparison.
	The attn1511 model furthermore has only $N/2$ RNN memory units and $N/2$ CNN filters.
	\\
$^*$ Exact models from \cite{UbuntuLowe} reran on the v2 version of the dataset (by personal communication with Ryan Lowe) --- note that the results in \cite{UbuntuLowe} and \cite{UbuntuKadlec} are on dataset v1 and not directly comparable.
}
\label{tab:ubuntu}
\end{figure*}

\begin{figure*}[p]
\centering
\begin{tabular}{|c|ccccc|c|c|}
\hline
                & SICK-2014 & SICK-2014 & SICK-2014 & SNLI & SNLI \\
Model           & STS $r$   & 3-RTE & 3-RTE & 3-RTE & 3-RTE \\
                & test      & train & test & train & test \\
\hline
\cite{MaLSTM} & $0.882$ &   &  $0.842$ &  & \\
\cite{IllinoisLH} &  & $0.842$ & $0.845$ &  & \\
\cite{SNLI} LSTM &  & $1.000$ & $0.713$ & $0.848$ & $0.776$\\
\cite{SNLI} Tran. &  & $0.999$ & $0.808$ &  & \\
\cite{LSTMMR} &  &  &  & $0.921$ & $0.890$\\
\hline
\hline
TF-IDF & $0.479$ &  &  &  & \\
BM25 & $0.474$ &  &  &  & \\
avg & $0.621$ & $0.770$ & $0.652$ & $0.735$ & $0.710$\\
 & $\quad^{\pm0.017}$ & $\quad^{\pm0.020}$ & $\quad^{\pm0.017}$ & $\quad^{\pm0.014}$ & $\quad^{\pm0.008}$\\
DAN & $0.642$ & $0.715$ & $0.662$ & $0.718$ & $0.708$\\
 & $\quad^{\pm0.016}$ & $\quad^{\pm0.010}$ & $\quad^{\pm0.003}$ & $\quad^{\pm0.009}$ & $\quad^{\pm0.002}$\\
\hline
RNN & $0.664$ & $0.759$ & $0.732$ & $0.784$ & $0.749$\\
 & $\quad^{\pm0.022}$ & $\quad^{\pm0.016}$ & $\quad^{\pm0.010}$ & $\quad^{\pm0.019}$ & $\quad^{\pm0.010}$\\
CNN & $0.762$ & $0.927$ & $0.799$ &  & \\
 & $\quad^{\pm0.006}$ & $\quad^{\pm0.008}$ & $\quad^{\pm0.004}$ &  & \\
RNN-CNN & \textbf{0.790} & $0.765$ & $0.709$ & $0.811$ & $0.753$\\
 & $\quad^{\pm0.005}$ & $\quad^{\pm0.084}$ & $\quad^{\pm0.059}$ & $\quad^{\pm0.037}$ & $\quad^{\pm0.008}$\\
attn1511 & $0.723$ & $0.858$ & $0.767$ & $0.829$ & \textbf{0.774}\\
 & $\quad^{\pm0.009}$ & $\quad^{\pm0.010}$ & $\quad^{\pm0.004}$ & $\quad^{\pm0.014}$ & $\quad^{\pm0.004}$\\
\hline
Ubu. RNN & \textbf{0.799} & $0.931$ & $0.813$ &  & \\
 & $\quad^{\pm0.009}$ & $\quad^{\pm0.017}$ & $\quad^{\pm0.005}$ &  & \\
SNLI RNN & \textbf{0.798} & $0.927$ & \textbf{0.831} &  & \\
 & $\quad^{\pm0.007}$ & $\quad^{\pm0.006}$ & $\quad^{\pm0.002}$ &  & \\
\hline
\end{tabular}
\vspace*{-0.2cm}
\caption{\footnotesize%
	Model results on the STS and RTE tasks, reporting Pearson's $r$ and 3-class accuracy, respectively.
	The \textbf{SNLI Tran.} baseline transfers SNLI-learned weights to the SICK-2014 task.
}
\label{tab:stsent}
\end{figure*}

\subsection{Results}

In Fig.~\ref{tab:anssel} to~\ref{tab:stsent}, we show the cross-task
performance of our models.  We can observe an effect analogous to what
has been described in \cite{UbuntuKadlec} --- when the dataset is smaller,
CNN models are preferrable, while larger dataset allows RNN models to capture
the text comprehension task better.  IR baselines provide strong competition
and finding new ways to ensemble them with models should prove beneficial
in the future.%
\footnote{We have tried simple averaging of predictions (as per \cite{UbuntuKadlec}), but the benefit
was small and inconsistent.}
This is especially apparent in the new Answer Sentence Selection datasets
that have very large number of sentence candidates per question.
The attention mechanism also has the highest impact in this kind of Information
Retrieval task.

On the Ubuntu Dialog dataset, even the simple LSTM model
in our proposed setting beats the baseline performance reported by Lowe,
while our RNN-CNN model establishes the new state-of-art,
beating the three-hop Memory Network of \cite{E2EDPrereq}.
It is not possible to statistically determine the relation of our models
to state-of-art
on the \texttt{wang} Answer Sentence Selection dataset.
Our models clearly yet lag behind the state-of-art on the RTE
and STS tasks, where we did not carefully tune their parameters,
but also did not employ data augmentation strategies like synonyme substitution
in \cite{MaLSTM}, which might be necessary for good performance on small
datasets even when using transfer learning.

\section{Model Reusability}
\label{sec:transfer}

To confirm the hypothesis that our models learn a generic task akin to some
form of text comprehension, we trained a model on the large Ubuntu Dialogue
dataset (Next Utterance Ranking task) and transferred the weights and retrained
the model instance on other tasks.
We used the RNN model for the experiment in a configuration with dot-product scorer
and smaller dimensionality (which works much better on the Ubuntu dataset).
This configuration is shown in the respective result tables as \textbf{Ubu.\ RNN}
and it consistently ranks as the best or among the best classifiers,
dramatically outperforing the baseline RNN model.%
\footnote{The RNN configuration used for the transfer, when trained only on the target task,
is not shown in the tables but has always been worse than the baseline RNN configuration.}

During our experiments,
we have noticed that it is important not to apply dropout during re-training
if it wasn't applied during the source model training, to balance the
dataset labels, and we used the RMSprop training procedure since Adam's learning
rate annealing schedule might not be appropriate for weight re-training.
We have also tried freezing the weights of some layers, but this never yielded
a significant improvement.

\cite{SNLI} have shown that such a model transfer is beneficial by reusing
an RTE model trained on the SNLI dataset to the SICK-2014 dataset.  We have
tried the same, shown as \textbf{SNLI RNN}, and while we see an improvement
when reusing it on an RTE task, on other tasks it is the same or worse than
the Ubuntu Dialogue based transfer, possibly because the Ubu.\ task sees more
versatile and less clean data.

\section{Conclusion}
\label{sec:concl}

We have unified a variety of tasks in a single scientific framework of sentence
pair scoring, and demonstrated a platform for general modelling of this problem
and aggregate benchmarking of these models across many datasets.  Promising initial
transfer learning results suggest that a quest for generic neural model capable
of task-independent text comprehension is becoming a meaningful pursuit.
The open source nature of our framework and the implementation choice of a popular and extensible
deep learning library allows for high reusability of our research and easy extensions with further more advanced models.

Based on our benchmarks, as a primary model for applications on new $f_2$-type tasks, we can
recommend either the RNN-CNN model or transfer learning based on the Ubu.\ RNN model.

\subsection{Future Work}

Due to the very wide scope of the $f_2$-problem scope, we leave some popular tasks and datasets
as future work.
A popular instance of sentence pair scoring is the question answering task of the
\textbf{Memory Networks} (supported by the baBi dataset) \cite{baBi}.
A realistic large question \textbf{Paraphrasing} dataset based on the AskUbuntu Stack Overflow
forum had been recently proposed \cite{AskUbuntu}.%
\footnote{The task resembles paraphrasing, but is evaluated as an Information Retrieval
task much closer to Answer Sentence Selection.}
In a multi-lingual context, sentence-level MT Quality Estimation is a meta-task with
several available datasets.%
\footnote{\url{http://www.statmt.org/wmt15/quality-estimation-task.html}}
While the tasks of \textbf{Semantic Textual Similarity} (supported by a dataset from\
the STS track of the SemEval conferences \cite{STS2015})
and \textbf{Paraphrasing} (based on the Microsoft Research Paraphrase Corpus \cite{MSRPara} right now)
are available within our framework, we do not report the results here
as the models lag behind the state-of-art significantly and show little difference
in results.  Advancing the models to be competitive remains future work.
A generalization of our proposed architecture could be applied to the
\textbf{Hypothesis Evidencing} task of binary classification of a hypothesis sentence $s_0$ based
on a number of memory sentences $s_1$, for example within the MCText \cite{MCTest} dataset.
%
%
We also did not include several major classes of models in our initial evaluation.
Most notably, this includes serial RNNs with attention as used e.g.\ for the RTE
task \cite{SNLIattn}, and the skip-thoughts method of sentence embedding. \cite{skipthoughts}

We believe that the Ubuntu Dialogue Dataset results demonstrate that the time
is ripe to push the research models further towards the real-world by allowing
for wider sentence variability and less explicit supervision.  But in particular,
we believe that new models should be developed and tested on tasks with long
sentences and wide vocabulary.
%
In terms of models, recent work in many NLP domains \cite{attnpooling} \cite{LSTMMR} \cite{DMN}
clearly points towards various forms of attention modelling to remove the
bottleneck of having to compress the full spectrum of semantics into a single
vector of fixed dimensionality.
%
In this paper, we have shown the benefit of training a model on a single dataset
and then applying it on another dataset.  One open question is whether we could
jointly train a model on multiple tasks simultaneously (even if they do not share
some output layers).
Another option would be to include extra supervision similar
to the token overlap features that we already employ; for example, in the new
Answer Sentence Selection task datasets, we can explicitly mark the actual
tokens representing the answer.

\section*{Acknowledgments}
{\footnotesize
This work was financially supported by the Grant Agency of the Czech Technical
University in Prague, grant No. SGS16/ 084/OHK3/1T/13, and the Augur Project of the Forecast Foundation.
Computational resources were provided by the CESNET LM2015042 and the CERIT Scientific Cloud LM2015085,
provided under the programme ``Projects of Large Research, Development, and Innovations Infrastructures.''

We'd like to thank Tom\'{a}\v{s} Tunys, Rudolf Kadlec, Ryan Lowe, Cicero Nogueira dos santos and Bowen Zhou for helpful discussions and their insights,
and Silvestr Stanko and Ji\v{r}\'{i} N\'{a}dvorn\'{i}k for their software contributions.}

\bibliography{sps}
\bibliographystyle{acl2016}

\end{document}